\documentclass[conference]{IEEEtran}
\IEEEoverridecommandlockouts
\usepackage{cite}
\usepackage{amsmath,amssymb,amsfonts}
\usepackage{algorithmic}
\usepackage{graphicx}
\usepackage{textcomp}
\usepackage[colorlinks,
            linkcolor=red,
            anchorcolor=blue,
            citecolor=green
            ]{hyperref}
\usepackage{xcolor}
\def\BibTeX{{\rm B\kern-.05em{\sc i\kern-.025em b}\kern-.08em
    T\kern-.1667em\lower.7ex\hbox{E}\kern-.125emX}}
\usepackage{multirow}
\usepackage{tikz}
\usepackage{booktabs}

\definecolor{yellowtext}{RGB}{68,132,243}
\definecolor{yellowred}{RGB}{50,167,82}
\definecolor{yellowblue}{RGB}{251,191,5}

\newcommand{\TextCircle}[1][0.7]{%
    \tikz[baseline=(char.base)]\node[shape=circle,draw=black,inner sep=1.5pt,line width=0.5pt,fill=yellowtext,text=white,scale=#1] (char) {T};\hspace{-1pt}
}
\newcommand{\ImageCircle}[1][0.76]{%
    \tikz[baseline=(char.base)]\node[shape=circle,draw=black,inner sep=1.5pt,line width=0.5pt,fill=yellowred,text=white,scale=#1] (char) {I};\hspace{-1pt}
}
\newcommand{\MultiCircle}[1][0.66]{%
    \tikz[baseline=(char.base)]\node[shape=circle,draw=black,inner sep=1.5pt,line width=0.5pt,fill=yellowblue,text=white,scale=#1] (char) {M};\hspace{-1pt}
}

\begin{document}

\title{Why Text Prevails: Vision May Undermine Multimodal Medical Decision Making
\thanks{This study was partially supported by the NIH (U01AG068057), the NSF (CCF 2523787, and IIS 2045848), as well as by the Presidential Research Fellowship (PRF) in the Department of Computer Science at the University of Texas Rio Grande Valley (UTRGV).}
}

\author{Siyuan Dai\IEEEauthorrefmark{1}, Lunxiao Li\IEEEauthorrefmark{2}, Kun Zhao\IEEEauthorrefmark{1}, Eardi Lila\IEEEauthorrefmark{3},  Paul K. Crane\IEEEauthorrefmark{4}, \\ Heng Huang\IEEEauthorrefmark{5}, Dongkuan Xu\IEEEauthorrefmark{2}, Haoteng Tang\IEEEauthorrefmark{6}, Liang Zhan\IEEEauthorrefmark{1}\IEEEauthorrefmark{7} \\
\IEEEauthorrefmark{1} Department of Electrical \& Computer Engineering, University of Pittsburgh, Pittsburgh, PA, USA \\
\IEEEauthorrefmark{2} Department of Computer Science, NC State University, Raleigh, NC, USA\\
\IEEEauthorrefmark{3} Department of Biostatistics, University of Washington, Seattle, WA, USA\\
\IEEEauthorrefmark{4} Department of Medicine, University of Washington, Seattle, WA, USA \\
\IEEEauthorrefmark{5} Department of Computer Science, University of Maryland, College Park, MD, USA \\
\IEEEauthorrefmark{6} Department of Computer Science, University of Texas Rio Grande Valley, Edinburg, TX, USA \\
\IEEEauthorrefmark{7} Corresponding author\\
\href{mailto:SID51@pitt.edu}{siyuan.dai@pitt.edu}, \href{mailto:liang.zhan@pitt.edu}{liang.zhan@pitt.edu}}


\maketitle

\begin{abstract}
With the rapid progress of large language models (LLMs), advanced multimodal large language models (MLLMs) have demonstrated impressive zero-shot capabilities on vision–language tasks. In the biomedical domain, however, even state-of-the-art MLLMs struggle with basic Medical Decision Making (MDM) tasks. We investigate this limitation using two challenging datasets: (1) three-stage Alzheimer’s disease (AD) classification (normal, mild cognitive impairment, dementia), where category differences are visually subtle, and (2) MIMIC-CXR chest radiograph classification with 14 non–mutually exclusive conditions. Our empirical study shows that text-only reasoning consistently outperforms vision-only or vision–text settings, with multimodal inputs often performing worse than text alone. To mitigate this, we explore three strategies: (1) in-context learning with reason-annotated exemplars, (2) vision captioning followed by text-only inference, and (3) few-shot fine-tuning of the vision tower with classification supervision. These findings reveal that current MLLMs lack grounded visual understanding and point to promising directions for improving multimodal decision making in healthcare.
\end{abstract}


\section{Introduction}
Building on the exceptional capabilities of large language models (LLMs), multimodal large language models (MLLMs) have emerged as promising tools for general medical AI~\cite{zhao2023evaluating}, offering broad zero-shot performance across diverse vision–language tasks~\cite{achiam2023gpt,dai2025zeus, li2023llava}. By jointly processing visual and textual inputs, these models extend the reasoning and comprehension abilities of LLMs to medical imaging domains such as radiology~\cite{thawakar2024xraygpt,zhao2025dre,zhao2024x,xiao2025overview}, pathology \cite{sun2025cpath}, and multi-omics~\cite{cui2024scgpt}.

Currently, increasing attention is being directed toward developing agentic MLLMs~\cite{kim2024mdagents,schick2023toolformer,li2024mmedagent,zhao2024slide} to tackle more sophisticated benchmarks. However, their effectiveness fundamentally depends on the ability of external tools to accurately interpret out-of-text information. This raises a critical question:
\textit{Do current MLLMs truly possess a grounded understanding of visual patterns?}

\begin{figure}
\includegraphics[width=0.5\textwidth]{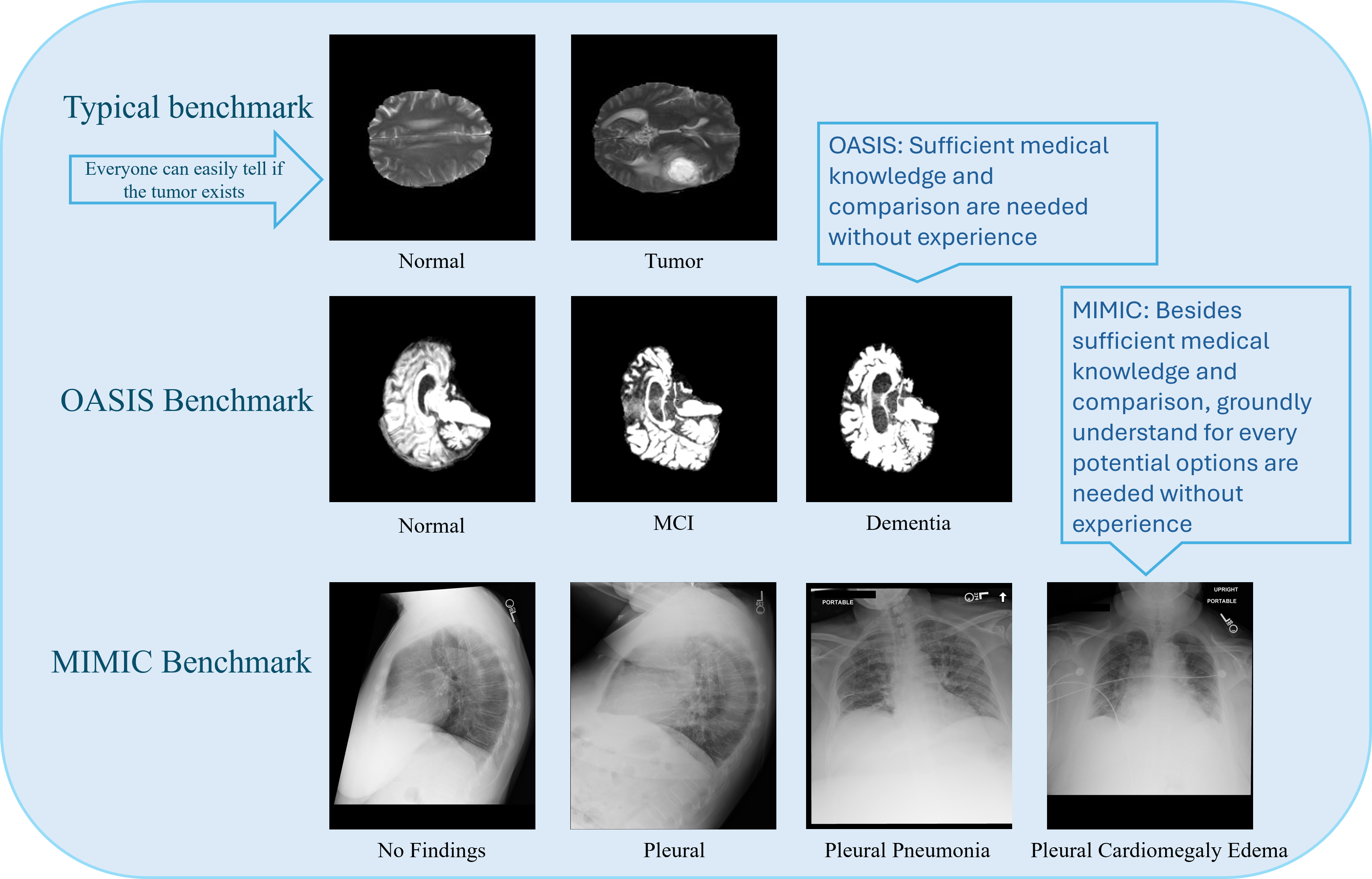}
\caption{Comparison among data from typical medical image analysis benchmark (row 1), our used OASIS (row2) and MIMIC-CXR (row3) benchark, highlighting the main gap in current Medical Decision Making tasks.} \label{fig1}
\end{figure}

As illustrated in Fig.~\ref{fig1}, typical medical decision-making (MDM) benchmarks are relatively easy to solve because the answer options often exhibit explicit and distinguishable visual patterns ~\cite{fang2022refuge2,guo2024soullmate1,zhao2023mitea,guo2024soullmate2,dai2025sin,dai2024constrained,tang2024ex}. 
In such cases, MLLMs may succeed by merely detecting anomalies (e.g., tumors, lesions) rather than demonstrating a grounded understanding of underlying semantics. To address this limitation, we focus on two more representative and challenging tasks:
1.	Three-class Alzheimer’s disease (AD) classification (normal, mild cognitive impairment, dementia), where the visual differences among categories are subtle \cite{tang2024interpretable,guo2023investigating,tang2025instantaneous,ye2025bpen,ye2023bidirectional}. Correct classification requires a holistic understanding of the entire brain; without domain knowledge and expertise, even human observers struggle to distinguish these classes.
2.	MIMIC-CXR, a multi-label chest X-ray diagnosis task covering 14 thoracic conditions and abnormalities. Because these pathologies are not mutually exclusive and often co-occur within the same image, the task requires a grounded understanding of the complete image and integration of relevant medical knowledge.

Both tasks require nuanced visual reasoning beyond simple anomaly detection. However, our empirical study reveals a critical phenomenon: when textual features alone are sufficient for predicting the correct label, MLLMs often achieve their best performance in the text-only setting. In contrast, the vision–text setting frequently underperforms compared to text alone, and the vision-only setting shows substantial accuracy degradation. These results suggest that, in such scenarios, the visual modality not only fails to contribute additional value but may even impair overall performance.

\begin{table*}[t]
\centering
\scriptsize
\caption{Evaluations of MLLMs and CLIP-based classifiers on our two MDM benchmarks. \TextCircle is text-only settings, \ImageCircle represents image-only settings, and \MultiCircle for multimodal vision-text settings.}
\setlength{\tabcolsep}{7mm}
{
\begin{tabular}{clcccccc}
\hline
\multirow{2}*{\textbf{Category}} &\multirow{2}*{\textbf{Models}} & \multicolumn{3}{c}{OASIS} & \multicolumn{3}{c}{MIMIC-CXR} \\ 
{} & {} & \TextCircle & \ImageCircle & \MultiCircle & \TextCircle & \ImageCircle & \MultiCircle  \\ \hline

\multirow{2}*{Contrastive Model} & CLIP & {-} & {43.23} & {-} & {-} & {0} & {-}  \\ 
{} & BiomedCLIP  & {-} & {35.34} & {-} & {-} & {9.00} & {-} \\  \hline

\multirow{4}*{Public MLLM} & LLaVA-Med & {7.31} & {21.05} & {7.78} & {6.50} & {0} & {3.50} \\
{} & LLaVA-Next & {14.66} & {12.78} & {14.29} & {32.00} & {4.00} & {23.50} \\
{} & InstructBLIP2 & {79.26} & {4.98} & {8.82} & {15.00} & {43.50} & {20.00} \\
{} & XrayGPT & {27.64} & {34.42} & {27.15} & {5.50} & {15.00} & {8.00} \\ \hline

\multirow{2}*{Proprietary MLLM} & GPT-4o & {80.07} & {12.78} & {83.08} & {46.00} & {38.00} & {41.00} \\ 
{} & Gemni2.5 Pro & {34.62} & {34.96} & {44.36} & {31.50} & {30.00} & {29.50} \\ \hline 

\multirow{5}*{MLLM Agent(s)} & +CoT & {81.20} & {5.64} & {80.83} & {0} & {1.00} & {0} \\ 
{} & +CoT-SC & {47.74} & {3.01} & {69.92} & {43.50} & {26.00} & {40.00}  \\
{} & +Debating & {57.14} & {5.26} & {79.70} & {59.00} & {42.50} & {16.00} \\
{} & MedAgents & {84.44} & {19.17} & {84.21} & {42.00} & {38.00} & {24.50}  \\ 
{} & MDAgents & {79.70} & {15.41} & {73.68} & {54.00} & {2.50} & {45.00}  \\ \hline
\end{tabular}
}
\label{tab:main}
\end{table*}

We conduct an empirical study with state-of-the-art MLLMs to investigate the underlying causes of this behavior and to explore strategies for mitigation. As an initial enhancement, we propose using in-context learning, where exemplars with ground truth are provided to guide the models, thereby improving zero-shot inference. We also outline other promising directions for future solutions. Overall, our results offer new insights into the limitations of current MLLMs on MDM benchmarks, highlighting that existing foundation models do not yet guarantee robust, grounded visual understanding.

\section{Experimental Setup}
\subsection{Foundation Models}
To deeply investigate the capabilities and limitations of modern multimodal large language models (MLLMs) in Medical Decision Making (MDM), we experiment with both open-source and closed-source foundational MLLMs with multi-categories. Besides analyzing how proprietary and public MLLMs, we also use two CLIP-based state-of-the-art vision-language classifiers \cite{radford2021learning,zhang2023biomedclip} as comparison. For every public MLLM is finetuned based on the CLIP model \cite{radford2021learning}, so they should theoretically have better classification capacity than the original foundation model. 

\subsubsection{Close-source MLLMs}
We evaluate two leading close-source LLMs with multimodal reasoning ability: GPT-4o \cite{achiam2023gpt} and Gemini-2.5 Pro \cite{comanici2025gemini}. Both are the most influential LLMs for their strong performance in general-purpose multimodal reasoning and zero-shot capabilities in visual understanding. In addition, numerous studies have explored advanced performance with established agentic models \cite{tang2024medagents,kim2024mdagents} based on these models, e.g., role-play, voting, debating, etc., without finetuning. In our study, we use 5 influential single-agent and multi-agent models as extra baselines, exploring their ability in our more challenging benchmarks. CoT \cite{wei2022chain} can always be recognized as one of the most single-agent methods for improving the ability without further training. CoT-SC is another influential single-agent model to enhance performance based on a few-shot strategy. Debating \cite{du2023improving} is a typical multi-agent method and also a fundamental multi-agent strategy used by other multi-agent models like MedAgents \cite{tang2024medagents} and MDAgents \cite{kim2024mdagents}. Additionally, all of the agentic models are based on GPT-4o.

\subsubsection{Open-source MLLMs}
We also include four state-of-the-art open-source MLLMs: LLaVA-Med \cite{li2023llava}, LLaVA-Next \cite{liu2024improved}, InstructBLIP \cite{dai2023instructblip}, and XrayGPT \cite{thawakar2024xraygpt}, which represent different vision–language integration strategies and training tricks. LLaVA-Med and LLaVA-Next are the most LLaVA-based \cite{liu2023visual} models finetuned in the medical domain and general domain, respectively. InstructBLIP and XrayGPT follow the structure from BLIP-2 \cite{li2023blip}. InstructBLIP is the most powerful BLIP-based MLLM, and XRayGPT is finetuned with one of our benchmark datasets, MIMIC-CXR \cite{johnson2019mimic}. These models all demonstrate strong generalizability and zero-shot inference ability on other multimodal benchmarks, including QA, captioning, classification, etc.

\subsection{Datasets}
To expose the limitations of MLLMs in grounded medical vision tasks, we select two challenging and underexplored datasets with contrasting properties: one neuroimaging and one chest X-ray dataset. Each requires careful comprehension of visual features and detailed reasoning, rather than simple pattern recognition or anomaly detection.

\subsubsection{OASIS}
We used the OASIS-3 \cite{lamontagne2019oasis} dataset for a three-class classification task based on dementia: Normal Control (NC), Mild Cognitive Impairment (MCI), and Dementia. It is inherently challenging due to the subtlety of the visual patterns. Unlike tasks with clear lesions, dementia-related changes require a holistic interpretation of brain atrophy across regions, involving subtle volumetric differences \cite{tang2022hierarchical,tang2023comprehensive,ye2025bpen,yin2024heterogeneous}.

\subsubsection{MIMIC-CXR}
The MIMIC-CXR dataset originally contains over 370,000 chest radiographs with associated diagnostic reports. It is a multi-label chest X-ray classification task involving 14 thoracic conditions or abnormalities. This benchmark with the pathologies can co-occur and often exhibit overlapping. This complexity requires a grounded understanding of the entire image with medical knowledge. In our study, we use the refined version of the dataset from XrayGPT \cite{thawakar2024xraygpt}, and for every test, we randomly select 200 samples for inference. It is worth mentioning that, for CLIP-based classifiers \cite{radford2021learning,zhang2023biomedclip},  we regard it as a true prediction if the prediction exists in the ground-truth label for the predicted one-hot results. However, for those generative models, the prediction should be 100\% aligned with the ground-truth labels.

\begin{table*}[t]
\centering
\scriptsize
\caption{In-context learning strategy for enhancing the performance of typical baselines. OASIS benchmark sees significant improvement, especially on image-only setting. Effectiveness is relatively slight on MIMIC-CXR dataset but show great potential when it is implemented with Agent0-based methods.}
\setlength{\tabcolsep}{6mm}
{
\begin{tabular}{clcccccc}
\hline
\multirow{2}*{\textbf{Category}} &\multirow{2}*{\textbf{Models}} & \multicolumn{3}{c}{OASIS} & \multicolumn{3}{c}{MIMIC-CXR} \\ 
{} & {} & \TextCircle & \ImageCircle & \MultiCircle & \TextCircle & \ImageCircle & \MultiCircle  \\ \hline

\multirow{4}*{Proprietary MLLM} & GPT-4o & {80.07} & {12.78} & {83.08} & {46.00} & {38.00} & {41.00} \\ 
& w/ enhancement & {68.50} & {57.29} & {84.24} & {42.00} & {24.50} & {32.00} \\
{} & Gemni2.5 Pro & {34.62} & {34.96} & {44.36} & {31.50} & {30.00} & {29.50} \\ 
& w/ enhancement & {37.26} & {35.13} & {49.91} & {32.00} & {24.00} & {28.00} \\\hline 

\multirow{4}*{MLLM Agent(s)} & CoT & {81.20} & {5.64} & {80.83} & {0} & {1.00} & {0} \\ 
{} & w/ enhancement & {84.00} & {8.03} & {87.13} & {6.00} & {4.50} & {6.50}  \\
{} & Debating & {57.14} & {5.26} & {79.70} & {59.00} & {42.50} & {16.00} \\
{} & w/ enhancement & {72.08} & {14.39} & {82.56} & {48.50} & {45.00} & {22.50}  \\ \hline
\end{tabular}
}
\label{tab:enc}
\end{table*}

\subsection{Evaluation Details}
We design three complementary evaluation settings to disentangle the contributions of vision and text modalities in MDM tasks. The results of decision accuracy is attached in Table \ref{tab:main}.

\subsubsection{Text Setting}
In the text-only setting, no image is provided. Instead, for the OASIS benchmark, we transfer the psychological and cognitive test results to text reports as input, while using the refined findings from X-ray images designed for XrayGPT training as input in the MIMIC-CXR benchmark. This setup simulates scenarios where textual summaries (e.g., from clinicians or external captioners) are available, and tests whether the model can accurately predict based on those descriptions. 

\subsubsection{Vision Setting}
In the vision-only setting, we remove all knowledge-related textual prompts and input only the image.  As image data in OASIS are 3D MRI images and the MLLMs are not designed to process 3D format, we select the middle three slices as visual input. And for both benchmarks, we resize all the input images to a $224*224$ shape for convenience.

\subsubsection{Multimodal Setting}
In the multimodal setting, the MLLMs receive both the image in the same format in the vision setting, and a templated textual prompt, identity to the text setting. 

\section{Empirical Study}
\subsection{Zero-shot Inference}
To validate the multimodal reasoning capability of MLLMs in MDM tasks and explore the vision effect in the whole prediction, we systematically evaluate the predicted accuracy of all baseline models across three modalities: vision-only ($\ImageCircle$), text-only ($\TextCircle$), and multimodal (text-vision $\MultiCircle$). The results are summarized in Table~\ref{tab:main}, spanning contrastive models, public and proprietary MLLMs, and MLLM agents with enhanced reasoning capabilities.

\subsubsection{Text-only setting outperforms others} 
Across nearly all models and both datasets, the text-only setting achieves the highest performance. For example, MedAgents reaches 84.44\% on OASIS stands the best performance, and GPT-4o with CoT performs the second with the accuracy of 81.20\%. While for MIMIC-CXR in the text-only setting, GPT-4o with a debating agentic setting conduct 59.00\% for the highest accuracy, and MDAgents reaches 54.00\%, showing the second-best results. Multi-agent models can easily beat the single public MLLMs or proprietary MLLMs. InstructBLIP shows a close result in the OASIS dataset with 79.26\% accuracy, while LLaVA-Next is the best public MLLM in the MIMIC-CXR benchmark.

\subsubsection{Vision-only setting suffers degradation}
Vision-only performance is mostly the lowest across all models, with many results approaching random baselines (e.g., 0–4\% for MIMIC-CXR). When it comes to the OASIS dataset, every baseline is randomly guessed, with a lower than 34.96\% by Gemini2.5 pro and close to the random selection for a three-category classification task. Even BiomedCLIP, pretrained on the largest medical image-text pairs, performs poorly when isolated from textual cues to 35.34\%. For the MIMIC-CXR benchmark, the result is very poor even for XrayGPT, which conducts in-distribution inference. This underscores that current visual encoders, even domain-adapted ones, cannot perform fine-grained medical classification independently.

\subsubsection{Multimodal fusion cannot help}
As for the multimodal intention for every research, people hope that different features in multimodalities can help each other, and most of the alignment mechanisms can show an increase in other easy benchmarks. However, according to our results, the poorer modality input often leads to performance degradation instead of complementing the other one. For instance, InstructBLIP2 achieves 79.26\% on OASIS (text-only), but drops sharply to 4.98\% when vision–text is provided. GPT-4o with debating follows a similar pattern in MIMIC-CXR (59.00\% $\rightarrow$16.00\%). Only GPT-4o, Gemini2.5-Pro show improvement in OASIS, while nothing shows any help in MIMIC-CXR. This suggests a misalignment between the multimodal features, especially when models only have poor grounding capability.

\subsection{In-context Learning}
To mitigate the issue and better guide the reasoning process of MLLMs, we introduce an \textbf{in-context learning} (ICL) strategy. Specifically, in each evaluation setting, we randomly select three labeled samples from different categories and incorporate them directly into the query prompt as exemplars. Each exemplar consists of the input (text, vision, or multimodal description) paired with its ground-truth label. The inference query is appended after these exemplars, making the examples part of the context window. Since public MLLMs cannot process multiple images within a single forward pass, we only conduct this evaluation with proprietary MLLMs (GPT-4o, Gemini-2.5 Pro) and agent-based methods (CoT and Debating). The results are illustrated on Table \ref{tab:enc}.

\subsubsection{Results on OASIS}  
The in-context learning strategy yields dramatic improvements for OASIS benchmark classification, especially under the vision setting. GPT-4o, for instance, improves from 12.78\% (vision–text zero-shot) to 57.29\% when enhanced with ICL. Similarly, the Debating agent sees an increase from 5.26\% to 14.29\%. These results highlight that even a small number of exemplars can substantially boost performance by forcing the model to reason from different categories, improving the ground-level comprehension ability.

\subsubsection{Results on MIMIC-CXR}  
The gains on the multi-label chest X-ray task are slight, and in some cases negative for single MLLMs. We attribute this to two factors: (1) only three exemplars are provided, which is insufficient to cover the full space of 14 categories. (2) The dataset is heavily imbalanced, leading to poor exemplar coverage of minority classes. Despite these challenges, agent-based methods such as CoT and Debating still demonstrate significant improvements under ICL, especially in vision and multimodal settings, suggesting the potential of comparing and groundly reasoning with vision-centric tasks.

\section{Limitation and Conclusion}
In this work, we systematically investigate the roles of text and vision modalities in multimodal medical decision making (MDM) with current MLLMs. Comprehensive experiments on OASIS and MIMIC-CXR show that text-only reasoning consistently dominates, vision–text fusion often degrades performance, and vision-only models remain inadequate. To bridge this gap, we explore strategies such as in-context learning to enhance inference-time reasoning, improving benchmark performance—particularly for agent-based methods—thereby highlighting the promise of contextual reasoning.
We also acknowledge limitations. Different MDM tasks demand varying levels of visual understanding, and future work should establish a principled ranking of task difficulty to guide the development of adaptive agentic models, while exploring broader strategies for multimodal enhancement.

\section{Acknowledgments}
Part of the work utilized Bridges-2 \cite{brown2021bridges} at the Pittsburgh Supercomputing Center through the ACCESS program, supported by NSF grants \#2138259, \#2138286, \#2138307, \#2137603, and \#2138296.

We also thank the providers of the \textbf{OASIS} and \textbf{MIMIC-CXR} datasets. 
For OASIS, the magnetic resonance imaging and neuropsychological test data that support the findings of this study are available in [``OASIS-3"] (\url{https://doi.org/10.1101/2019.12.13.19014902}). OASIS-3: Longitudinal Multimodal Neuroimaging: Principal Investigators: T. Benzinger, D. Marcus, J. Morris; NIH grants P30 AG066444, P50 AG00561, P30 NS09857781, P01 AG026276, P01 AG030991, R01 AG043434, UL1 TR000448, R01 EB009352. AV-45 doses were provided by Avid Radiopharmaceuticals, a wholly owned subsidiary of Eli Lilly.
OASIS-3 data\footnote{\url{https://www.oasisbrains.org}} are openly available to the scientific community. Prior to accessing the data, users are required to agree to the OASIS Data Use Terms (DUT), which follow the Creative Commons Attribution 4.0 License.

\bibliographystyle{IEEEtran}
\bibliography{mybib}

\end{document}